\documentclass{article}

\usepackage{times}
\usepackage{graphicx} 
\usepackage{subfigure} 

\usepackage{natbib}

\usepackage{algorithm}
\usepackage{algorithmic}

\usepackage{hyperref}

\usepackage[accepted]{icml2016}

\usepackage{url}
\usepackage{latexsym}

\usepackage{times}
\usepackage{amsmath}
\usepackage{amsthm}
\usepackage{amssymb}
\usepackage{xspace}
\usepackage{tabularx}
\usepackage{multicol}
\usepackage{multirow}
\usepackage{url}
\usepackage{wrapfig}
\usepackage{comment}
\usepackage{listings}
\usepackage{color}
\usepackage[utf8]{inputenc}
\usepackage{fancyvrb}
\usepackage{booktabs}
\usepackage{color}
\usepackage[normalem]{ulem}

\newcommand{\bmx}[0]{\begin{bmatrix}}
\newcommand{\emx}[0]{\end{bmatrix}}

\newcommand{\vect}[1]{\mathbf{#1}}

\newcommand{\va}[0]{\vect{a}}
\newcommand{\vb}[0]{\vect{b}}

\newcommand{\vh}[0]{\vect{h}}

\newcommand{\vx}[0]{\vect{x}}

\newcommand{\vw}[0]{\vect{w}}

\newcommand{\vm}[0]{\vect{m}}

\newcommand{\MM}[0]{\mathcal{M}}
\newcommand{\NN}[0]{\mathcal{N}}

\icmltitlerunning{A Controller-Recognizer Framework}

\begin{document} 

\twocolumn[
\icmltitle{A Controller-Recognizer Framework: \\ How necessary is recognition for control?}

\icmlauthor{Marcin Moczulski}{marcin.moczulski@stcatz.ox.ac.uk}
\icmladdress{University of Oxford}
\icmlauthor{Kelvin Xu}{kelvin.xu@umontreal.ca}
\icmladdress{Universit\'{e} de Montr\'{e}al}
\icmlauthor{Aaron Courville}{aaron.courville@umontreal.ca}
\icmladdress{Universit\'{e} de Montr\'{e}al}
\icmlauthor{Kyunghyun Cho}{kyunghyun.cho@nyu.edu}
\icmladdress{New York University}

\icmlkeywords{boring formatting information, machine learning, ICML}

\vskip 0.3in
]

\begin{abstract} 
    Recently there has been growing interest in building ``active'' visual
    object recognizers, as opposed to ``passive'' recognizers which classifies
    a given static image into a predefined set of object categories. In this
    paper we propose to generalize recent end-to-end active visual recognizers
    into a controller-recognizer framework. In this framework, the interfaces
    with an external manipulator, while the recognizer classifies the visual
    input adjusted by the manipulator. We describe two recently proposed
    controller-recognizer models-- the recurrent attention
    model \cite{mnih2014recurrent} and spatial transformer
    network \cite{jaderberg2015spatial}-- as representative examples of
    controller-recognizer models. Based on this description we observe that
    most existing end-to-end controller-recognizers tightly couple the
    controller and recognizer. We consider whether this tight coupling is
    necessary, and try to answer this empirically by investigating a decoupled
    controller and recognizer. Our experiments revealed that it is not always
    necessary to tightly couple them, and that by decoupling the controller and
    recognizer, there is a possibility to build a generic controller that is
    pretrained and works together with any subsequent recognizer.
\end{abstract} 

\section{Introduction}
\label{sec:intro}
The success of deep learning, in particular convolutional networks, in computer
vision has largely been due to breakthroughs in {\em passive} object
recognition from a {\em static} image
\citep{krizhevsky2012imagenet,lecun1998gradient}. Most of the successful
convolutional networks for object recognition
\citep{szegedy2014going,simonyan2014very} are passive in the sense that they
recognize an object without having any ability to act on it to improve
recognition. Also, they work with static images in the sense that these models
lack the ability or mechanism to manipulate an input image themselves.

It has been only very recently that these passive neural network {\em
recognizers} have become more active. This is often done by letting the model
actively attend to a sequence of smaller regions of an input image
\citep{mnih2014recurrent,ba2014multiple,Denil2012a} or by allowing the model to
distort the input image \citep{jaderberg2015spatial}. In general, these
recognizers have become {\em active} by allowing them access to a controller
which acts upon an external manipulator that either adjusts the recognizer's
view or controls an external mechanism that directly manipulates the
environment (which is viewed by the recognizer as an image.)

In this paper, we generalize this shift in the paradigm of object recognition
with neural networks by defining a {\em controller-recognizer framework}. In
this framework, a neural network based object recognition system consists of a
controller and a recognizer. The {\em recognizer} can be any object recognizer
that perceives the surrounding environment as a 2-D image.  The {\em
controller} has access to an external mechanism (often a black-box) which can
either adjust the surrounding environment directly or a view of the recognizer.
A full controller-recognizer model is defined by the exact specifications of
how these recognizer and controller components are coupled with each other.

We show that many recently proposed neural network based active object
recognizers fall into this controller-recognizer family. More specifically, we
explain in detail the specifications of the recurrent attention
model~\citep[RAM,][]{mnih2014recurrent} and the spatial transformer
network~\citep[STN,][]{jaderberg2015spatial} under this controller-recognizer
framework. From this we notice that these existing controller-recognizers
tightly, or sometimes completely, couple the controller and recognizer such
that the recognizer has deep access to the inner workings of the controller or
that the controller relies heavily on the recognizer. 

Based on this observation, in this paper we ask ourselves whether this is the
only option in designing an end-to-end trainable controller-recognizer model.
This question is natural considering the number of potentially undesirable
properties of a tightly-coupled pair of controller and recognizer such as 
a lack of a clear way to use existing well-performing recognizer architectures
(i.e.  convolutional classifiers) or limited compatibility with external
black-box manipulators.

As a first stab at answering this question, we design a controller-recognizer
model with a decoupled controller/recognizer. In this decoupled model, the
controller first manipulates an input image by issuing a sequence of image
manipulation commands to an external, non-differentiable image manipulator.
After a fixed number of commands were issued, the resulting image is sent to
the recognizer for it to detect an object. In this setting, the internal
representations of the controller and recognizer are completely separate.

With this decoupled controller-recognizer model, we test a wide variety of
training strategies to empirically confirm (1) the possibility of training a
decoupled model jointly and (2) the possibility of having a general, pretrained
controller for a subsequent recognition task with potential mismatch between
the data used to train the controller and a recognizer. Furthermore, we aim to
show that the existing benchmark task of recognizing a randomly placed
handwritten digit in a large canvas (potentially with clutter) can in fact be
solved by this decoupled model at a level comparable to a tightly coupled
model.

Our experimental results show that a  decoupled controller-recognizer model
achieves a level of performance comparable with a tightly coupled model and performs
well in transfer settings. This opens the potential of having a model with a
{\em single, generic controller} manipulating the environment to maximize the
performance of {\em multiple recognizers}.

\section{Controller-Recognizer Framework}
\label{sec:crframework}

\begin{figure}[ht]
    \centering
        \centering
        \includegraphics[width=0.95\columnwidth]{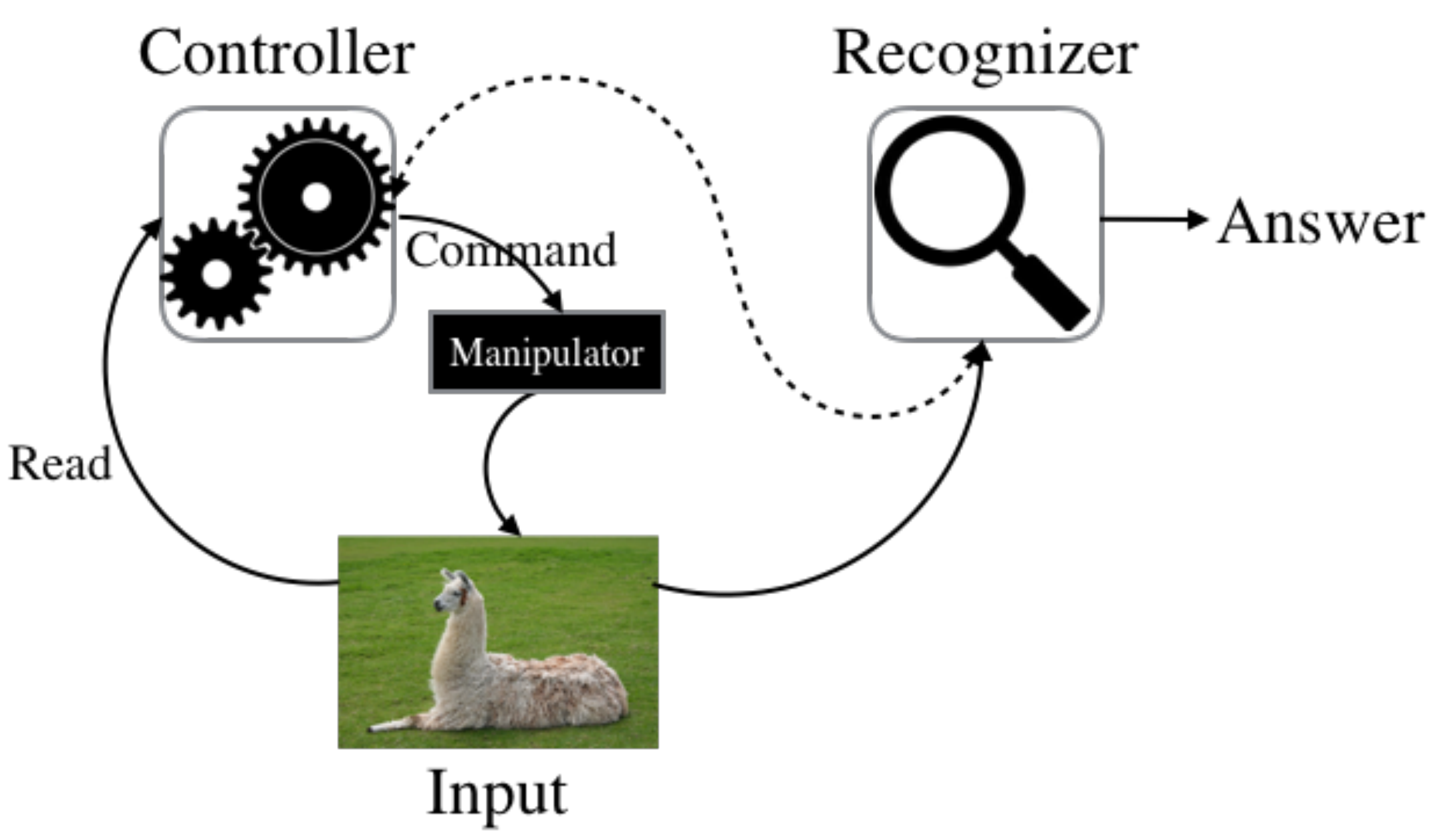}
        \caption{Graphical illustration of a general controller-recognizer
        model. Solid arrows indicate the flow of information, and a dashed arrow
    is an optional information path.}
        \label{fig:controller_recognizer}
\end{figure}

In this paper, we are interested in models that exploit the ability of control
in order to recognize an object based on vision. These models can be described
as consisting of a {\em controller} and a {\em recognizer}. In general, a
controller of this type of models manipulates either the external environment
or the model itself to adjust the model's view. This adjusted view of the
external environment is used by a recognizer, and therefore the controller's
objective is to adjust the view so as to maximize the recognizer's performance.
See Fig.~\ref{fig:controller_recognizer} for a graphical illustration.

This is in contrast to existing supervised object recognition models such as
the widely used convolutional neural network \citep[see,
e.g.,][]{lecun1998gradient,krizhevsky2012imagenet}. This approach to
vision-based object recognition using neural networks is {\em static} approach
in the sense that the model does have any means of influencing the environment.
This means that the model has to work with whatever input, but has no control
over how it can be manipulated in order to maximize recognition rate.

\subsection{Criteria for Categorizing Controller-Recognizer Models}
\label{sec:criteria}

Our {\em controller-recognizer} framework includes a broad family of models.
Among these we are interested in fully-trainable end-to-end models, often
implemented as a deep neural network. Although we focus on a
subset of models, comprised of end-to-end trainable deep neural networks, there
are many possible variants, either already proposed or not, and in this
section, we try to describe how they can be categorized.

First, we can classify each of these controller-recognizer models based on the
{\em type of manipulator} used by the controller. A controller may manipulate
the model itself in order to move its gaze over a static input.\footnote{ By
static input we mean a situation when a model does not actively, directly
manipulate the observed environment.} In recent literature, this is often
referred to as an attention mechanism \citep[see,
e.g.,][]{Denil2012a,zheng2014neural,mnih2014recurrent,ba2014multiple}. On the
other hand, a controller may also have access to an external, often black-box,
mechanism, and this external black-box manipulates the input actively based
upon the commands issued by the controller. 

Second, a controller-recognizer model can be classified based upon the {\em
training objective(s) of the controller}. The ultimate goal of the controller
is eventually to maximize the recognition rate by the recognizer, but this does
not necessarily imply that this is the only training objective available. For
instance, a controller may be trained jointly to focus on an object of interest
as well as to explore the input scene (i.e., maximize the model's coverage over
the input scene) in order to detect the existence of an object which will
ultimately be recognized. In this case, the training objectives for the
controller are (1) to maximize the recognition rate and (2) to maximize the
exploration.

Another closely-related criterion is the level of {\em generality of the
controller}. By the generality of the controller, we mean specifically whether
a given controller can be used for multiple recognition tasks. As evident in
animals, a single controller can be utilized for multiple downstream
recognition tasks (visual recognition, speech recognition, haptic perception
etc.) One can use the controller to bring an object in interest to the center
of view to better recognize, or at the same time use it to remove distractions
in the scene (i.e., denoising.) 

Yet another criterion is specific to neural network based controller-recognizer
models. Regardless of its end goal, a deep neural network automatically extracts
a continuous vector representation of the input. A neural controller will have a
continuous vector representation of the original input, adjusted input (by
itself) and potentially a sequence of control commands. This representation may
be used by a recognizer, rather than having the recognizer work directly on the
adjusted input returned as a result of the controller.

This criterion reflects the {\em strength of coupling} between the controller
and the recognizer, and is closely related to the generality of the controller.
Stronger coupling implies that the controller's behaviour as well as its
internal representation are highly customized for a subsequent recognizer,
leading to less generality of the controller. On the other hand, when the
controller and recognizer are weakly coupled, the generality of the controller
increases and may be more suitable to be used with multiple training objectives
and recognition models. Therefore, we consider the generality of the controller
as a sub-criterion of the coupling strength.

The coupling strength also has consequences on the {\em training strategy}. If
the controller and recognizer are strongly coupled, it is quite likely that they
will have to be trained simultaneously. This is not necessary true if the
coupling strength is weak. In this case, one can think of bootstrapping, or
pretraining, the controller with another training objective, which is useful for
a wide set of potential downstream recognition tasks, before coupling this
pretrained controller with other recognizers.

We summarize this criteria here as a list, and next describe representative
examples under this framework:
\begin{enumerate}
    \itemsep 0em
    \item Manipulator: attention mechanism, external black-box, internal
        white-box, etc.
    \item Training objectives: final recognition rate, exploration rate, etc.
    \item Strength of coupling: how tightly a controller and recognizer are
        coupled
        \begin{enumerate}
            \item Generality of controller: single recognition task vs. multiple
                recognition tasks
            \item Training strategy: sequential vs. simultaneous
        \end{enumerate}
\end{enumerate}

\subsection{Example 1: Recurrent Attention Model}
\label{sec:ram}

A recurrent attention model (RAM) is a representative example of
controller-recognizer models, recently proposed by \citet{mnih2014recurrent}.
RAM was designed to work on a large image efficiently by controlling the
model's gaze of a small view area over the input image. 

At each time step, RAM receives as input a subset of the whole input
image from the gaze's location determined at the previous step. This subset is
used to update the hidden state (continuous vector representation of the input.)
Based on this updated state RAM computes the next location of the gaze,
which is equivalent to adjusting itself to move its gaze to the next location.
RAM also predicts when to stop and finally what is the object type.

\begin{figure}[ht]
    \centering
        \centering
        \includegraphics[width=0.8\columnwidth]{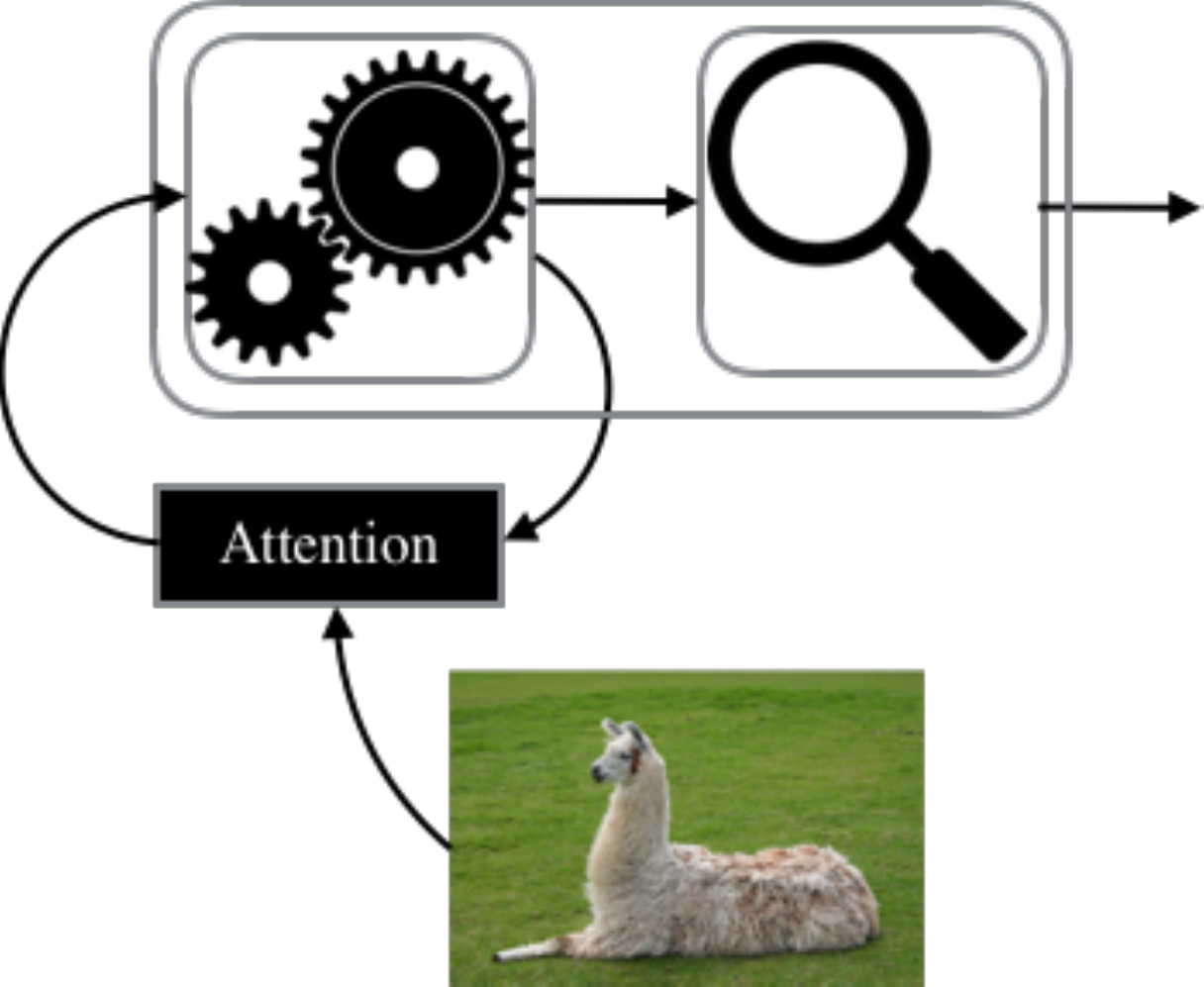}
        \caption{Graphical illustration of a recurrent attention model by
            \cite{mnih2014recurrent}. (1) Attention mechanism is used as a
        manipulator, (2) the controller is trained to maximize the recognition
    rate, and (3) the controller and recognizer are tightly coupled.}
            \label{fig:ram}
\end{figure}

We analyze this model according to the criteria we have defined earlier.
See Fig.~\ref{fig:ram} as a reference.

The manipulator used by the RAM is an {\em attention mechanism}.
\citet{mnih2014recurrent} however also showed that it is indeed possible to use
the RAM for playing a game, meaning that the RAM is able to interact with the
external black-box to maximise the final objective. 

Both the controller and the recognizer in RAM are tuned to maximize the {\em
final recognition rate} (or the reward from the game) which is the only
training objective.

The controller introduced as a part of the RAM is {\em tightly coupled} with
the recognizer by being a part of one recurrent neural network. The controller
and recognizer share the same set of parameters and the internal hidden state,
meaning that it is not possible to use the controller on another task once it
is trained together with the existing recognizer. This makes it difficult to
reuse the pretrained controller for another downstream recognition task, unless
all of them are trained simultaneously \citep[i.e., multitask
learning,][]{caruana1997multitask,collobert2011natural}.

\subsection{Example 2: Spatial Transformer Networks}
\label{sec:stn}

More recently, \citet{jaderberg2015spatial} proposed to modify a convolutional
neural network, which is a recognition only model, to include a controller. The
overall network is called a spatial transformer network (STN).

The STN employs a {\em differentiable spatial transformer} as a manipulator. The
spatial transformer is able to warp an input image based on transformation
parameters computed by a localisation network embedded inside a convolutional
neural network. The biggest advantage of having this differentiable spatial
transformer is that one can take the derivative of the final recognition rate
with respect to the transformation performed based on the transformation
parameters, which enables the use of backpropagation to compute a low-variance
gradient estimate.

\begin{figure}[ht]
    \centering
        \centering
        \includegraphics[width=0.7\columnwidth]{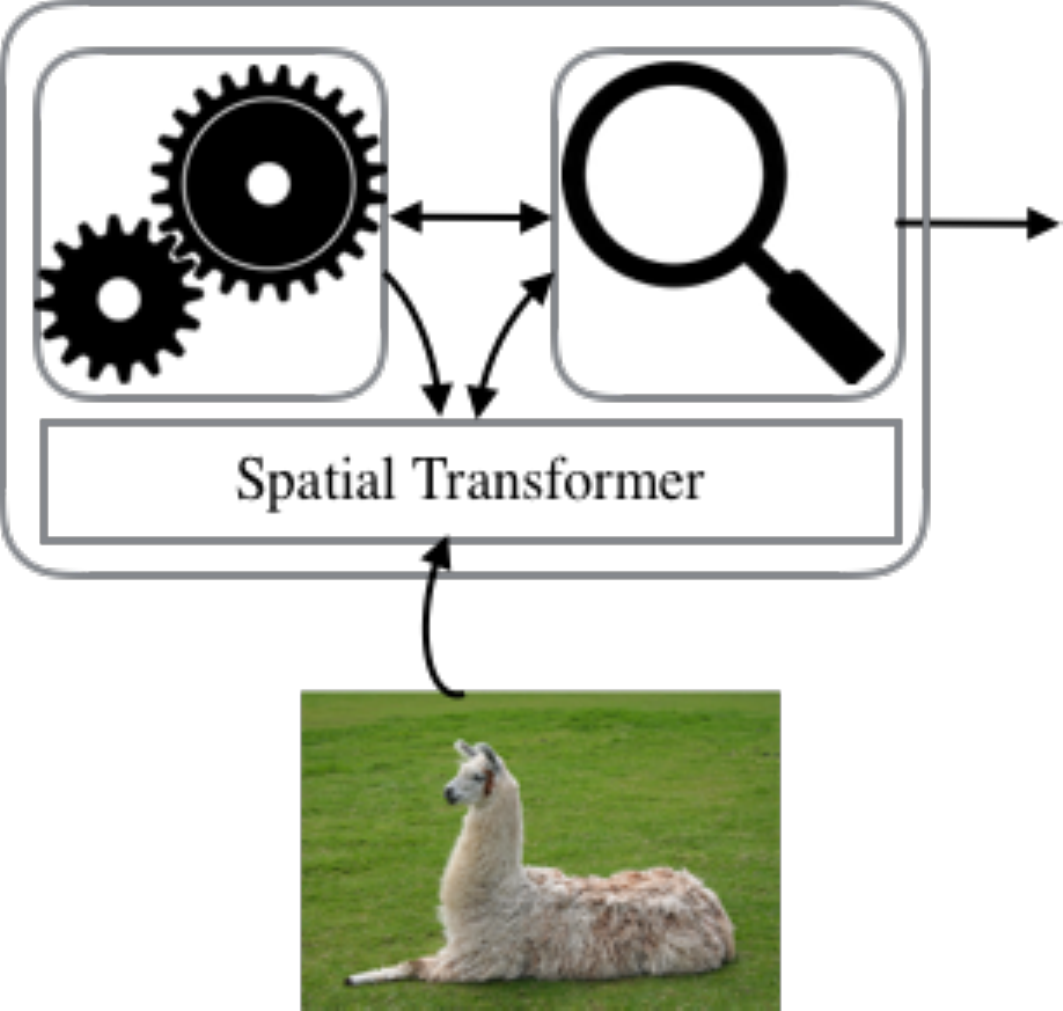}
        \caption{Graphical illustration of a spatial transformer network by
            \cite{jaderberg2015spatial}. (1) Spatial transformer is used as a
        white-box manipulator, (2) the controller is trained to maximize the recognition
    rate, and (3) the controller and recognizer as well as the manipulator are
completely coupled.}
            \label{fig:stn}
\end{figure}

Similarly to what we have done with the RAM, let us analyse the STN according to
the criteria for controller-recognizer models. See Fig.~\ref{fig:stn} as a
reference.

First, the target of the STN's controller is the {\em internal, transparent
manipulator}--the spatial transformer (ST). The ST works on either the raw input
image or the intermediate feature maps from the convolutional neural network. 
This manipulator, which works directly on the input image, is more flexible in
transforming the input than the attention mechanism of the RAM.

Second, one important characteristics of the STN is that both the
controller--localisation network-- and its target manipulator--spatial
transformer-- are all tuned to maximise the {\em final recognition rate}. 

As it is quite clear from its description, the controller and recognizer in the
STN are {\em completely coupled}. The controller works directly on the internal
continuous vector representation of the recognizer, and the recognizer's hidden
states are used as an input to the controller.  The controller, recognizer as
well as the manipulator {\em must be trained simultaneously}.

\section{Is It Necessary to Tightly Couple Controller and Recognizer?}

We noticed that both the recurrent attention model (RAM) and spatial transformer
network (STN) tightly, or completely, couple the controller and recognizer. This
is also observed in most of the recently proposed controller-recognizer models
such as the Fixation NADE by \citet{zheng2014neural}.

There are a number of implications from this tight coupling of the controller
and recognizer. 

This tight coupling, especially the complete coupling such as in the spatial
transformer network, implies that they need to be  trained simultaneously.
This simultaneous training naturally and obviously makes the controller
specialized for the recognition tasks used during training.  Consequently, it is
unclear whether the trained controller will be any useful for other subsequent
recognition tasks that arise after the original controller-recognizer is
trained. In other words, if there is another recognition task that may benefit
from having a controller, the whole new controller-recognizer will have to be
trained from scratch.

A further consequence of this reliance on a single objective of recognition rate
is that the controller of a controller-recognizer model can only be trained in a
supervised manner. This is unsatisfactory as the role of the controller is often
to bring an object to the center of the recognizer's view, which is
substantially a weaker, or easier, objective than the full recognition.

This reliance on the simultaneous training of the controller and recognizer is
quite contrary to what is observed in infant development. Infants are known to
exhibit visual attention already in the first few weeks after their birth
\citep[Chapter 3 of][]{ruff2001attention}. This happens without any strong
external reward, which is in the case of controller-recognizer framework a
recognition rate, implying that the controller, in this case an attention
mechanism similar to the one from the RAM, is being trained/tuned on its own.
This serves as an existence proof of the possibility of training a controller
separately from a recognizer also in this controller-recognizer framework.

\begin{figure}
    \centering
    \begin{minipage}{0.44\columnwidth}
        \centering
        \includegraphics[width=0.99\columnwidth]{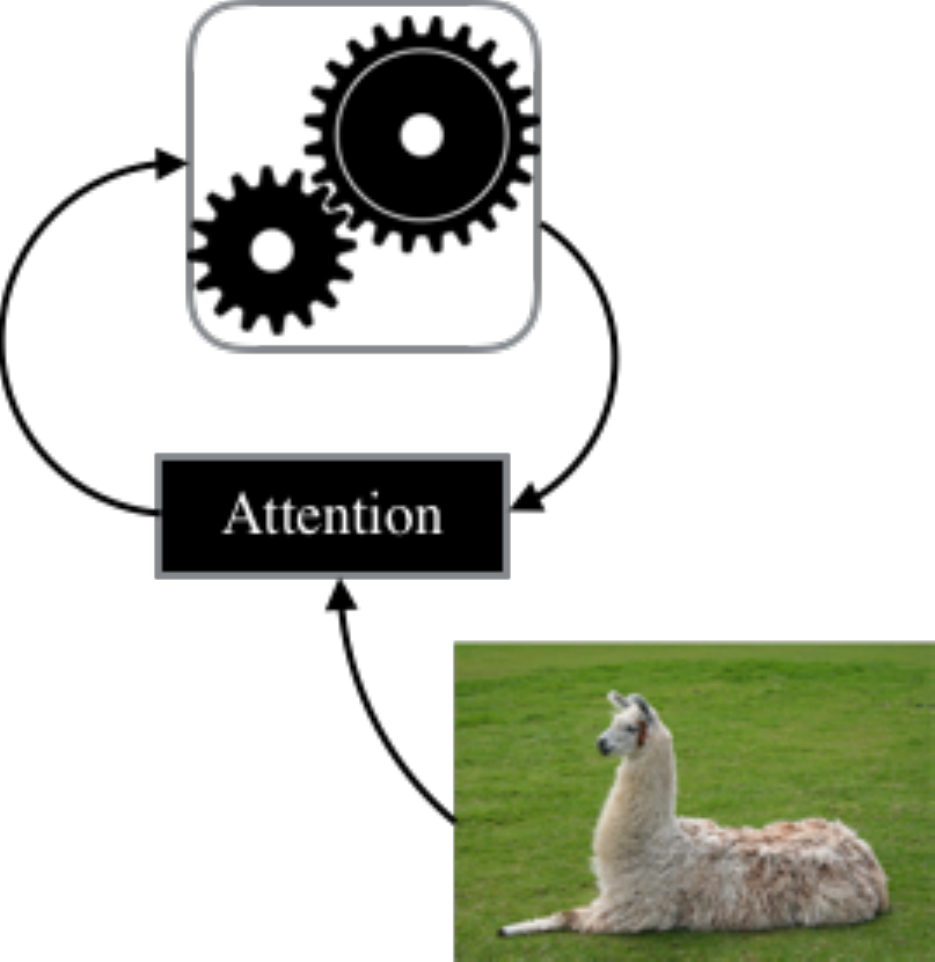}

        (a)
    \end{minipage}
    \hfill
    \begin{minipage}{0.55\columnwidth}
        \centering
        \includegraphics[width=0.99\columnwidth]{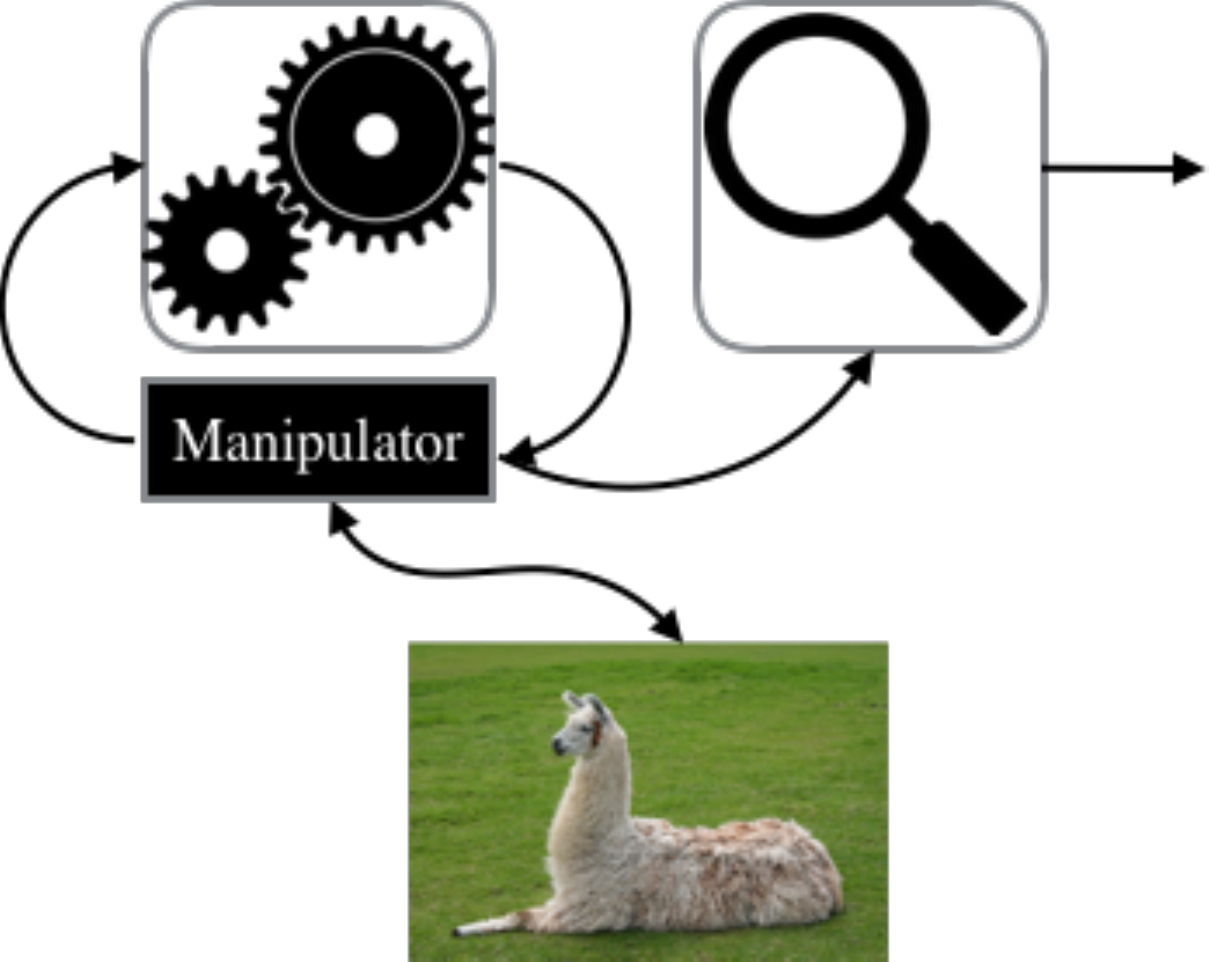}

        (b)
    \end{minipage}

    \caption{(a) Graphical illustration of a controller-only model by
        \cite{schmidhuber1991learning}. Note that there is no recognizer in
        this case. (b) A controller-recognizer model with decoupled controller
    and recognizer.}
    \label{fig:schmidhuber}
\end{figure}

Earlier in 1991, \citet{schmidhuber1991learning} proposed a very specific
approach to training a controller on its own without a subsequent recognizer, as
in Fig.~\ref{fig:schmidhuber}~(a).  Similarly to the RAM discussed earlier in
Sec.~\ref{sec:ram}, the controller in their case is an attention mechanism
implemented as a recurrent neural network, but without any recognizer. Their
goal was to show that it is possible to train a controller--attention mechanism
without explicit supervision on the types of objects the controller is
following. They achieved this by training the controller with a reward given
only when the controller managed to move its attention to a part of the input
image that contains an object.

The significance of the work by \citet{schmidhuber1991learning} is that the
controller {\em pretrained} in their method can be used later with a separate
recognizer that takes as input only a small subset of the input image selected
by this controller. There are two advantages in this procedure. First, the
recognizer can be made substantially simpler as it does not need to be invariant
to translation or rotation, as this is handled by the pretrained controller.
Second, as the recognizer takes as input only a small subset of the input image,
computational efficiency of the recognizer greatly increases.

These observations on infant development and the earlier work by
\citet{schmidhuber1991learning} raise a question on the necessity of strongly
coupling a controller and a recognizer in the controller-recognizer framework.
The success of those recent controller-recognizer models, such as the recurrent
attention model and spatial transformer network, does not answer this question.
This question naturally leads us to ask what other competitive variants within
general controller-recognizer framework, according to the criteria outlines in
Sec.~\ref{sec:criteria}, are possible. 

\section{Decoupled Controller and Recognizer}

In this paper, we aim at answering the questions posed in the previous section.
Among them, the main question is the possibility and extent of building a
controller-recognizer model with decoupled controller and recognizer. First, let
us describe what we mean by ``decoupled''.

A recognizer decoupled from a controller takes as input an image manipulated by
the controller only. In other words, the recognizer does not have access to the
internal state of the controller. One obvious consequence of this is that a
recognizer can be trained on its own regardless of the state of the controller,
although the ability of the controller in manipulating an input image will
significantly influence the final recognition quality.  Similarly, a controller
decoupled from a recognizer works independently from the recognizer. See from
Fig.~\ref{fig:schmidhuber}~(b) that there is no direct path between the
controller and recognizer.

\subsection{Model Description}
\label{section:model_description}

\paragraph{Input Canvas}

The world is a large $w \times h$ canvas $X$. On the canvas, a number of
objects, including the target object, are placed. The controller and recognizer
have their own window of view into the world. The recognizer always observes the
canvas through the center window of size $w_r \times h_r$, while the controller
has two possible views. Similarly to the recognizer, the controller may
observe the canvas through the center window of size $w_c \times h_c$ ({\em
cropped view}), or the
controller may view the whole canvas but in a lower resolution of $w_c \times
h_c$ ({\em subsampled view}.) Additionally, we test the case where the
controller is allowed the full view of the canvas ({\em full view}.)

\paragraph{Manipulator}

In this paper, we use an external black-box manipulator $\MM$ which permits a
set of possible actions $\left\{ a^1, a^2, \ldots, a^{N_a} \right\}$ and a gain
knob $p$ which decides on the degree to which a selected action is performed.
The manipulator we used is implemented using an imaging library with the
following actions; (1) shift up, (2) shift down, (3) shift right, and (4) shift
left.  For these actions, the gain knob $p\in \left[0, 1\right]$ corresponds to
the percentage of a pre-specified maximum shift rounded to the nearest integer.
Additionally, the manipulator may decide not to act on the input canvas by
issuing a {\em no}-action. 

\paragraph{Controller}

The controller is implemented as a recurrent neural network. At each time step,
the controller looks at the current configuration of the input canvas and
updates its internal hidden state:
\begin{align*}
    \vh_t = \phi\left( \vh_{t-1}, \vx_{t} \right),
\end{align*}
where $\phi$ is a recurrent activation function such as long short-term memory
units~\citep[LSTM,][]{hochreiter1997long} and gated recurrent
units~\citep[GRU,][]{cho2014learning}. In our experiments, we use 40 GRU's to
implement $\phi$.

The internal hidden state is initialized
as a function of the initial input canvas $\vx_0$:
\begin{align*}
    \vh_0 = f_{\text{init}}(\vx_0),
\end{align*}
which is a small multilayer perceptron. 

Given a new hidden state $\vh_t$, the controller computes the action
distribution by
\begin{align*}
    p(a_t = a^j | \vx_{<t}) = 
    \frac{\exp\left( \vw_{a^j}^\top \vh_t \right)}
    {\sum_{j'=1}^{N_a} \exp\left( \vw_{a^j}^\top \vh_t \right)},
\end{align*}
where $\vw_{a^j}$ is a parameter vector for the $j$-th action. The controller
then either samples, or selects the most likely, action $\tilde{a}_t$ from this
distribution.

Similarly, the controller computes the gain distribution
\begin{align*}
    p_t | \vx_{<t} \sim \NN\left(m(\vh_t), s(\vh_t)^2 \right),
\end{align*}
where $\NN(m, s^2)$ is a normal distribution with mean $m$ and variance $s^2$,
and
\begin{align*}
    m(\vh_t) = \sigma(\vw_m^\top \vh_t), s(\vh_t) = \sigma(\vw_s^\top \vh_t)
\end{align*}
We choose the gain $\tilde{p}_t^0$ to be either a sample or the mean of this
distribution. The selected gain is further processed to lie between $0$ and $1$:
\begin{align*}
    \tilde{p}_t = \sigma(5.5 \cdot (\tilde{p}_t^0 - 0.5)).
\end{align*}

The selected action and associated gain are fed into the manipulator. The
manipulator adjusts the input canvas $X$ accordingly to result in the next time
step's view $\vx_t$. We let the controller manipulate the input canvas for at
most $T$ steps, and we write by
\begin{align}
    \label{eq:controller}
    \vx_T = f_{\text{cont}}(\vx_0, \tilde{a}_{1:T}, \tilde{p}_{1:T}) 
\end{align}
to represent this whole process. Note that the manipulator $\MM$ is included
in $f_{\text{cont}}$, and that $\tilde{a}_{1:T}$ and $\tilde{p}_{1:T}$
are the sampled action and gain variables.

\paragraph{Recognizer}

As the controller is completely decoupled from a recognizer, we can use any
existing off-the-shelf image recognizer. More specifically, we use a generic
convolutional neural network with the configuration presented in
Table~\ref{table:convnet}. The recognizer returns a class-conditional
probability distribution over the labels $p(y=k|\vx_T)$, 
where $\vx_T$ is the final canvas configuration from Eq.~\eqref{eq:controller}.

\begin{table}[th]
    \centering
    \begin{tabular}{c || c | c | c | c }
        Type & Size & Stride & Output Size & Activation\\
        \hline
        \hline
        Convolution & 5x5 & 1,1 & 28x28x32 & $\tanh$ \\
        Max Pooling & 3x3 & 3,3 & 9x9x32 & -- \\
        Convolution & 5x5 & 1,1 & 5x5x64 & $\tanh$ \\
        Max Pooling & 2x2 & 2,2 & 2x2x64 & -- \\
        Linear & -- & -- & 200  & $\tanh$ \\
        Linear & -- & -- & 10  & softmax \\
    \end{tabular}
    \caption{The network configuration of the recognizer. }
\label{table:convnet}
\end{table}

\subsection{Training Strategies}
\label{sec:trainingstrats}

\begin{figure*}[t]
    \begin{minipage}{0.33\textwidth}
        \centering
        \includegraphics[width=0.9\columnwidth]{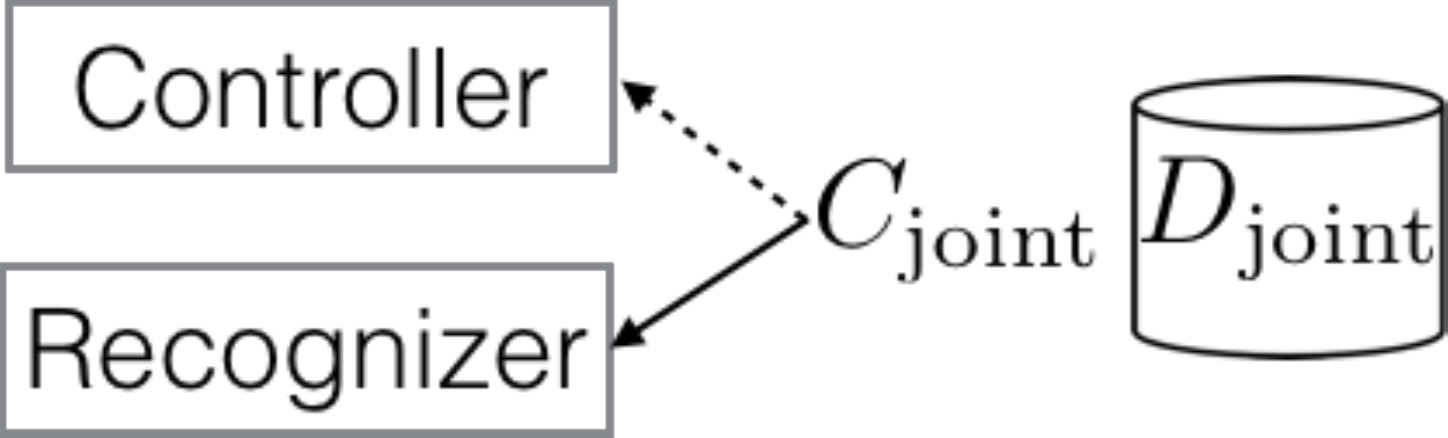}
    \end{minipage}
    \hfill
    \begin{minipage}{0.33\textwidth}
        \centering
        \includegraphics[width=0.9\columnwidth]{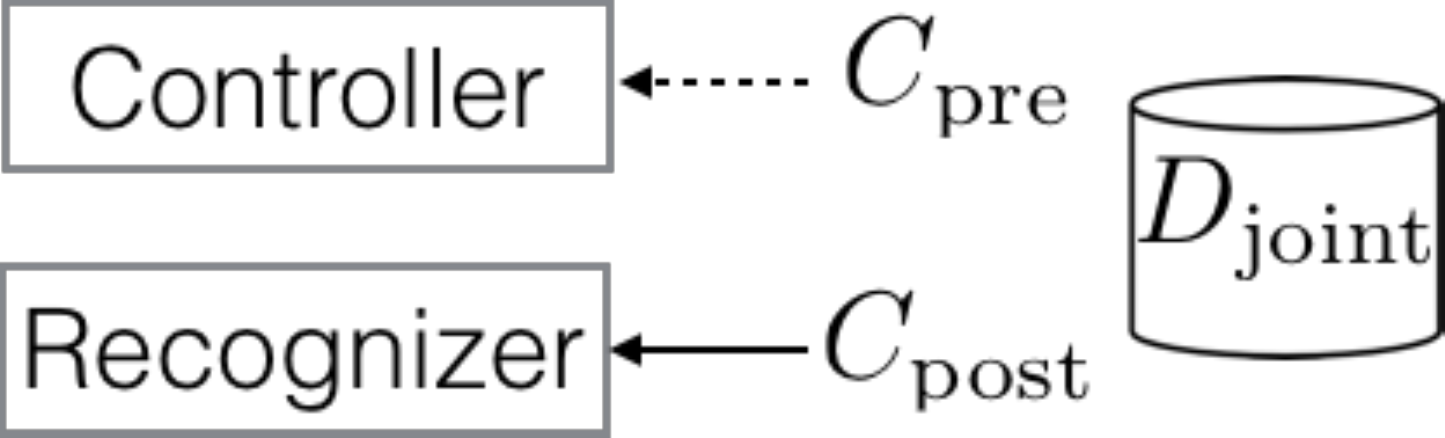}
    \end{minipage}
    \hfill
    \begin{minipage}{0.33\textwidth}
        \centering
        \includegraphics[width=0.9\columnwidth]{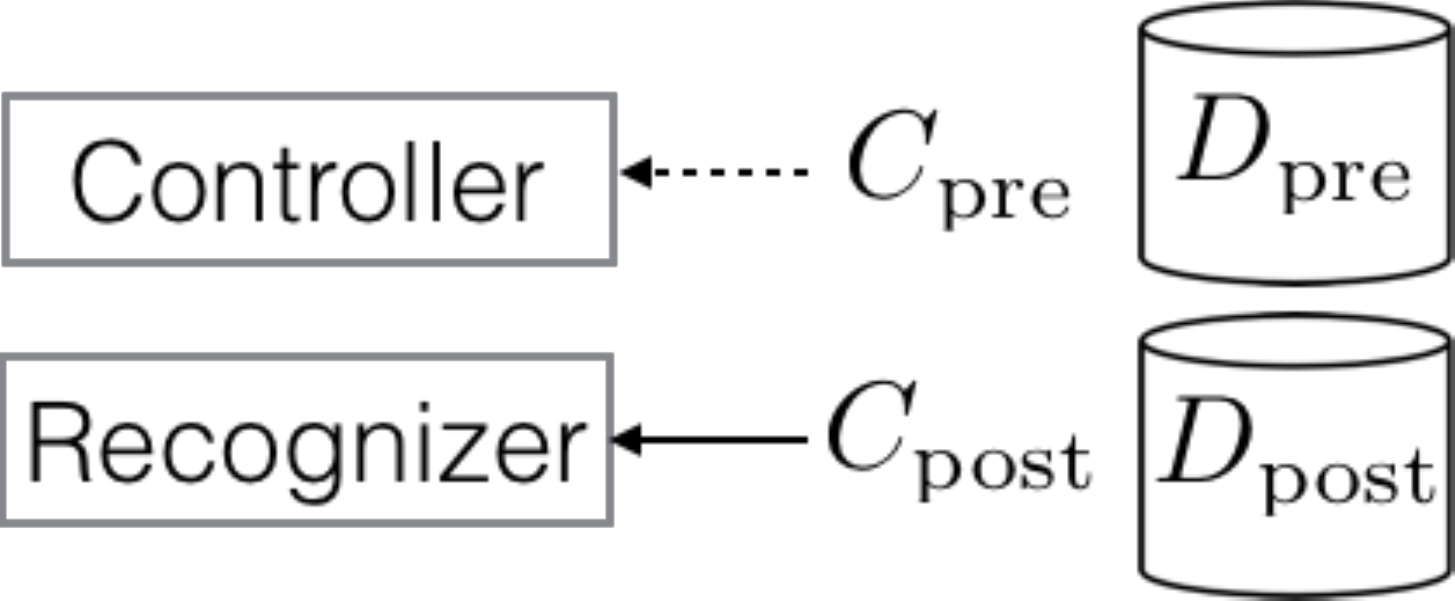}
    \end{minipage}

    \begin{minipage}{0.33\textwidth}
        \centering
        (a) Joint Training
    \end{minipage}
    \hfill
    \begin{minipage}{0.33\textwidth}
        \centering
        (b) Decoupled Training A
    \end{minipage}
    \hfill
    \begin{minipage}{0.33\textwidth}
        \centering
        (c) Decoupled Training B
    \end{minipage}

    \caption{Three training strategies. (a) Joint training. (b) Decoupled
        training with a single training set $D_{\text{joint}}$. (c) Decoupled
        training with separate training sets $D_{\text{pre}}$ and
    $D_{\text{post}}$. Solid arrows indicate training by backpropagation, and
dashed arrows by REINFORCE.}
    \label{fig:training}
\end{figure*}

\subsubsection{Joint Strategy}

Although the controller and recognizer are decoupled, we can jointly train them
to maximize 
\begin{align*}
    C_{\text{joint}} = \frac{1}{N}\sum_{n=1}^N c_{\text{joint}}(\vx_0^n, y^n),
\end{align*}
where 
\begin{align*}
    c_{\text{joint}}(\vx_0, y) = -\log \sum_{a_{1:T}, p_{1:T}} 
    p(y | f_{\text{cont}}(\vx_0, a_{1:T}, p_{1:T})).
\end{align*}
$(\vx_0^n, y^n)$ is the $n$-th example from a training set $D_{\text{joint}}$.

We use stochastic gradient descent (SGD) algorithm to update the parameters of
both the controller and recognizer, where the gradient $\nabla l$ is
approximated by
\begin{align*}
    \nabla c_{\text{joint}} \approx& -\frac{1}{M} \sum_{m=1}^M \nabla \log p(y |
    f_{\text{cont}}(\vx_0, a_{1:T}^m, p_{1:T}^m)) \\
    &+ \log p(y| f_{\text{cont}}(\vx_0, a_{1:T}^m, p_{1:T}^m)
    \left(
        \right.
    \\
    &
    \sum_{t=1}^T \nabla \log p(a_t=a_t^m|\vx_{<t})
    \\
    &+ \nabla \log p(p_t=p_t^m | \vx_{<t})
        \left.
        \right),
\end{align*}
where $a_{1:T}^m$ and $p_{1:T}^m$ are the action and gain variables sampled at
the $m$-th trial. In our experiment, we set $M=1$, meaning that we run the
controller only once for each training example. This approximation is necessary,
as the manipulator $\MM$ is non-differentiable.

This approximation to the gradient is known as
REINFORCE~\citep{williams1992simple}. We use REINFORCE to jointly train the
decoupled controller-recognizer model, augmented by the variance reduction
techniques proposed by \citet{mnih2014neural}.

We call this training approach the {\em joint strategy}.

\subsubsection{Decoupled Strategy}

Because the controller is separate from the recognizer, we can {\em pretrain} it
in advance of training the recognizer. We use the objective proposed by
\citet{schmidhuber1991learning}, where the controller's goal is to bring an
object in interest (which is known to the trainer) to the center of visual
perception which is in our case the center of the canvas. We define the cost of
each trial (i.e., running the controller for a single example) as
\begin{align*}
    c_{\text{pre}}(\vx_0, \vm) = 1-\text{cosine}(f_{\text{cont}}(\vx_0, \tilde{a}_{1:T},
    \tilde{p}_{1:T}),
    \vm),
\end{align*}
where $\vm$ is a mask vector corresponding to an input canvas with its $w'
\times h'$ center window set to 1 and 0 otherwise, and 
\begin{align*}
    \text{cosine}(\va, \vb) = \frac{\va^\top \vb}{\|\va\| \|\vb\|}.
\end{align*}
Similarly to joint training, we can minimize this pretraining cost function
by REINFORCE. Note that this pretraining does not require any labelled example
and can be done purely in a unsupervised manner. We denote by
the training set used to pretrain the controller as $D_{\text{pre}}$ .

\paragraph{Training Recognizer}
Once the controller is pretrained, we freeze it and train the recognizer.  This
is done, for each training example, by running the controller (and
manipulator) on the input canvas, feeding in the center window to the
recognizer, computing the gradient of the recognition cost, to which we refer as
$c_{\text{post}}$, w.r.t. the
recognizer's parameters and updating them accordingly: 
\begin{align*}
    \nabla c_{\text{post}} = \nabla \log p(y |
    f_{\text{cont}}(\vx_0, \tilde{a}_{1:T}, \tilde{p}_{1:T})).
\end{align*}

We use $D_{\text{post}}$ to refer to the training set used for tuning the
recognizer. Because the recognizer is trained separately from the controller, we
may either use the same training set, i.e., $D_{\text{post}}=D_{\text{pre}}$, or
a different set, i.e., $D_{\text{post}}\neq D_{\text{pre}}$. 

We call this training approach a {\em decoupled strategy}.  See
Fig.~\ref{fig:training}~(a)--(c) for graphical illustrations of these training
strategies. 

In addition, we also test a strategy where the decoupled strategy is followed
by finetuning the controller toward minimizing $C_{\text{post}}$, which we call
the {\em decoupled+finetune strategy}. 

\begin{table*}[th]
    \centering
    \begin{tabular}{c | c | c | c | c | c}
        & Controller's View & Training Strategy & 
        $D_{\text{pre}}$ & $D_{\text{joint}}$/$D_{\text{post}}$ & Error Rate (\%) \\
        \hline
        \hline
        (a) & Full & Joint & -- & CT-MNIST-Full & 3.82\% \\
        (b) & Full & Decoupled & CT-MNIST-Full & CT-MNIST-Full & 4.79\% \\
        (c) & Full & Decoupled+Finetune & CT-MNIST-Full & CT-MNIST-Full & 3.57\% \\
        (d) & Full & Decoupled & CT-MNIST-Thin & CT-MNIST-Thick & 2.68\% \\
        \hline
        (e) & Subsampled & Joint & -- & CT-MNIST-Full & 3.42\% \\
        (f) & Subsampled & Decoupled & CT-MNIST-Full & CT-MNIST-Full & 4.89\% \\
        (g) & Subsampled & Decoupled+Finetune & CT-MNIST-Full & CT-MNIST-Full & 3.53\% \\
        (h) & Subsampled & Decoupled & CT-MNIST-Thin & CT-MNIST-Thick & 2.28\% \\
        \hline
        (i) & Cropped & Joint & -- & CT-MNIST-Full & 4.28\% \\
        (j) & Cropped & Decoupled & CT-MNIST-Full & CT-MNIST-Full & 5.42\% \\
        (k) & Cropped & Decoupled+Finetune & CT-MNIST-Full & CT-MNIST-Full & 4.51\% \\
        (l) & Cropped & Decoupled & CT-MNIST-Thin & CT-MNIST-Thick & 3.32\% \\
        \hline
        (m) & Full & Decoupled & CT-MNIST-Full & CT-MNIST-Natural-$0.15$ & 4.24\% \\
        (n) & Full & Decoupled & CT-MNIST-Full & CT-MNIST-Natural-$0.25$ & 6.03\% \\
        \hline
        (o) & \multicolumn{4}{c|}{Recurrent Attention Model (RAM)$^\diamond$ } & 4.04\%$^\dagger$ \\
    \end{tabular}
    \caption{
        The results by all the combinations of controller's view (Full,
        Subsampled or Cropped), training strategy (Joint, Decoupled or
        Decoupled+Finetune), $D_{\text{pre}}$ and
        $D_{\text{post}}/D_{\text{joint}}$. See
        Sec.~\ref{section:model_description} for the detailed exposition of each
        column. ($\diamond$) The best performance
        reported by \citet{mnih2014recurrent}. $^\dagger$ Should be used for comparison primarily for CT-MNIST-Full experiments.
    }
    \label{tab:result1}

        \vspace{-4mm}
\end{table*}

\section{Experiments}
\label{sec:expsetting}

\subsection{Dataset}

We evaluate the proposed decoupled controller-recognizer model on the
classification task using the cluttered and translated MNIST (CT-MNIST), closely
following \citep{mnih2014recurrent} where the recurrent attention model (RAM,
see Sec.~\ref{sec:ram}) was proposed. 

Each example in the CT-MNIST consists of a $60\times 60$ canvas on which a
target handwritten digit together with multiple partial digits are randomly
scattered. The controller sees and manipulates the canvas by issuing commands to
the image library based manipulator. When the controller is done, the recognizer
looks at the $28\times 28$ center window of the final canvas and outputs the
label distribution. 

\begin{figure}[t]
\begin{minipage}{.49\columnwidth}
\begin{minipage}{.49\columnwidth}
\centering
    \includegraphics[width=\columnwidth]{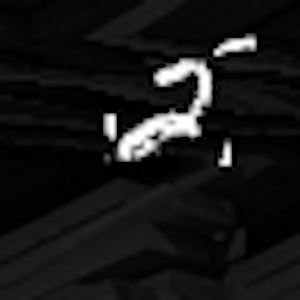}
\end{minipage}
\hfill
\begin{minipage}{.49\columnwidth}
\centering
    \includegraphics[width=\columnwidth]{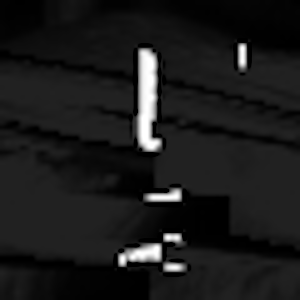}
\end{minipage}
\end{minipage}
\hfill
\begin{minipage}{.49\columnwidth}
    \begin{minipage}{.49\columnwidth}
    \centering
        \includegraphics[width=\columnwidth]{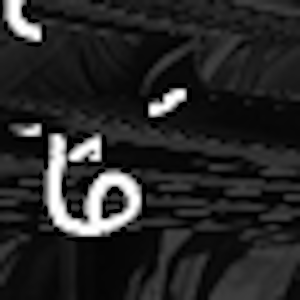}
    \end{minipage}
    \hfill
    \begin{minipage}{.49\columnwidth}
    \centering
        \includegraphics[width=\columnwidth]{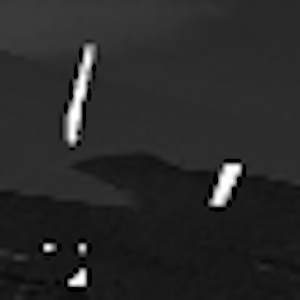}
    \end{minipage}
\end{minipage}

\begin{minipage}{.49\columnwidth}
    \centering
    (a)
\end{minipage}
\hfill
\begin{minipage}{.49\columnwidth}
    \centering
    (b)
\end{minipage}

\caption{Examples of (a) CT-MNIST-Natural-0.15 and (b)
CT-MNIST-Natural-0.25 (best viewed digitally)}
\label{fig:naturalmnist}
\end{figure}

We build the following subsets of the CT-MNIST for the decoupled training
strategies:
\begin{enumerate}
    \itemsep -.5em
    \item CT-MNIST-Full: CT-MNIST as it is
    \item CT-MNIST-Thin: Digits with labels $\left\{ 0, 1, 2, 3,
        9\right\}$
    \item CT-MNIST-Thick: Digits with labels $\left\{ 4, 5, 6, 7,
        8\right\}$
    \item CT-MNIST-Natural-$X$: CT-MNIST with natural image background\footnote{
            We use the Berkeley Segmentation Dataset \cite{MartinFTM01}.
        }
        of opacity set to $X$ (see Fig.~\ref{fig:naturalmnist})
\end{enumerate}
These datasets are used with the decoupled training strategy. By having
$D_{\text{pre}} \neq D_{\text{post}}$, we test the generality of the
pretrained controller.

\subsection{Results and Analysis}

In Table~\ref{tab:result1}, we report the recognition error rate on the test set
for each combination. We use the alphabet index to refer to a specific row
in the table.

\begin{figure}[t]
\begin{minipage}{.19\columnwidth}
\centering
    \includegraphics[width=\columnwidth]{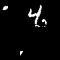}
\end{minipage}
\hfill
\begin{minipage}{.19\columnwidth}
\centering
    \includegraphics[width=\columnwidth]{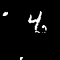}
\end{minipage}
\hfill
\begin{minipage}{.19\columnwidth}
\centering
    \includegraphics[width=\columnwidth]{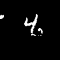}
\end{minipage}
\hfill
\begin{minipage}{.19\columnwidth}
\centering
    \includegraphics[width=\columnwidth]{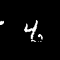}
\end{minipage}
\hfill
\begin{minipage}{.19\columnwidth}
\centering
    \includegraphics[width=\columnwidth]{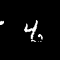}
\end{minipage}

\begin{minipage}{.19\columnwidth}
\centering
$\vx_0$
\end{minipage}
\hfill
\begin{minipage}{.19\columnwidth}
\centering
\end{minipage}
\hfill
\begin{minipage}{.19\columnwidth}
\centering
\end{minipage}
\hfill
\begin{minipage}{.19\columnwidth}
\centering
$\to$
\end{minipage}
\hfill
\begin{minipage}{.19\columnwidth}
\centering
\end{minipage}
\hfill
\begin{minipage}{.248\columnwidth}
\centering
$\vx_T$
\end{minipage}

\caption{Example sequence of manipulations by the decoupled controller.}
\label{fig:boat1}
\end{figure}

\paragraph{How well does the decoupled model do?}
From (a) and (o), we see that the proposed decoupled controller-recognizer
model, when jointly trained, works as well as the more tightly coupled
controller-recognizer model (RAM). However, we notice that when the controller
and recognizer are separately trained (row (b)), the performance slightly
degrades, but this is overcome by finetuning the controller subsequently (row
(c)). See Fig.~\ref{fig:boat1} for an example of the controller moving a digit
to the center window. This confirms that it is indeed possible to decouple the controller and
recognizer in the controller-recognizer framework. 

One potential criticism of the settings (a--c) is that the controller has the
full view of the canvas unlike the RAM of which controller has only a partial
view of the canvas, via attention mechanism, at a time. In the rows (e--g) and
(i--k), we present the results using the controller that has a more restricted
view of the canvas. In both cases, we see that the decoupled model works as well
as, or sometimes better than, the RAM, further supporting the decoupled model as
a viable model in the controller-recognizer framework.

\paragraph{How transferable is the pretrained controller?}
First, let us consider rows (d), (h) and (l). In these cases, the controller
was trained on CT-MNIST-Thin but was used for CT-MNIST-Thick. From the low
error rates, it is clear that the controller is able to easily manipulate the
digits that were not seen before for recognition. We further observed
qualitatively that this is indeed the case. This weakly supports that a
full-extent, fine-grained recognition is not necessary nor useful for control. 

Finally, from the rows (m--n) we see that the controller can manipulate the
canvas even when its background is covered with natural images which the
controller has never been exposed to before. As expected, the controller's
ability to manipulate degrades as the natural image background becomes
brighter, i.e., higher opacity, but despite the visible differences,
recognition performance degrades gracefully.

\section{Conclusion}

The main contribution of this paper is the introduction of a
controller-recognizer framework under which many recently proposed active
recognizers, such as recurrent attention model (RAM) and spatial transformer
network can be studied and analyzed. This framework allows us to view the active
recognizer as a composite of two separate modules, controller and recognizer,
and by doing so, gives us a systematic way to build a novel
controller-recognizer model and evaluate it. 

As an example, we proposed a decoupled controller-recognizer model, which
separates the controller and recognizer. This decoupling allows us to devise a
diverse set of learning and inference scenarios, such as pretraining a
controller on one data set and using it together with a recognizer on another
data set (transfer setting.) Our empirical evaluation confirms that the
proposed decoupled model indeed works well for most of these scenarios. These
experiments opens a door to a possibility of having a single, generic
controller is weakly coupled with a variety of subsequent recognizers. 

\subsubsection*{Acknowledgments}

The authors would like to thank the following for research funding and
computing support: NSERC, FRQNT, Calcul Qu\'{e}bec, Compute Canada, the Canada
Research Chairs, CIFAR and Samsung. Kyunghyun Cho and Kelvin Xu thank Facebook
for their generous support. The authors would also like to thank Çağlar
Gülçehre and Jamie Kiros for interesting discussions during the course of
this work.

\bibliography{manirec}
\bibliographystyle{icml2016}

\end{document}